\pdfoutput=1

\documentclass[11pt]{article}
\usepackage{EMNLP2024}
\usepackage{times}
\usepackage{latexsym}
\usepackage{multirow} 
\usepackage{booktabs}
\usepackage{amsmath}
\usepackage{array}
\usepackage{subfigure}
\usepackage{algorithmic}
\usepackage{algorithm}
\usepackage[T1]{fontenc}
\usepackage[utf8]{inputenc}
\usepackage{microtype}
\usepackage{graphicx}
\usepackage{inconsolata}
\usepackage{enumitem}
\usepackage{amssymb}
\usepackage{tcolorbox}
\usepackage{colortbl}
\usepackage{marvosym}
\newcommand{\nosection}[1]{\vspace{5.pt}\noindent\textbf{#1.}}

\title{SEER: Self-Aligned Evidence Extraction for Retrieval-Augmented Generation}

\author{
Xinping Zhao$^{1}$, Dongfang Li$^{1}$, Yan Zhong$^{2}$, Boren Hu$^{3}$, \\ \textbf{Yibin Chen$^{4}$,} \textbf{Baotian Hu$^{1}$\textsuperscript{\Letter}\thanks{\textsuperscript{\Letter}Corresponding author.},} \textbf{Min Zhang$^{1}$}  \\ 
$^1$Harbin Institute of Technology (Shenzhen), $^2$Peking University, \\
$^3$Zhejiang University, $^4$Huawei Cloud, Huawei Technologies Ltd. \\
\texttt{zhaoxinping@stu.hit.edu.cn}, \texttt{\{zhongyan946, huboren99\}@gmail.com}, \\ \texttt{chenyibin4@huawei.com}, \texttt{\{lidongfang, hubaotian, zhangmin2021\}@hit.edu.cn}
}

\makeatletter
\def\thanks#1{\protected@xdef\@thanks{\@thanks
        \protect\footnotetext{#1}}}
\makeatother

\begin{document}
\maketitle

\begin{abstract}
Recent studies in Retrieval-Augmented Generation (RAG) have investigated extracting evidence from retrieved passages to reduce computational costs and enhance the final RAG performance, yet it remains challenging.
Existing methods heavily rely on heuristic-based augmentation, encountering several issues: 
(1) Poor generalization due to hand-crafted context filtering; 
(2) Semantics deficiency due to rule-based context chunking; 
(3) Skewed length due to sentence-wise filter learning. 
To address these issues, we propose a model-based evidence extraction learning framework, \textbf{SEER}, optimizing a vanilla model as an evidence extractor with desired properties through self-aligned learning. 
Extensive experiments show that our method largely improves the final RAG performance, enhances the faithfulness, helpfulness, and conciseness of the extracted evidence, and reduces the evidence length by 9.25 times. The code will be available at \url{https://github.com/HITsz-TMG/SEER}.

\end{abstract}

\section{Introduction}
\label{sec:intro}
Recent years have witnessed the prevailing winds of Retrieval-augmented Generation (RAG), which is a popular paradigm for improving the performances of Large Language Models (LLMs) in various downstream tasks, 
such as question answering, making the output more reliable \cite{lewis2020retrieval,chen2023complex,jiang2023active,ram2023context}, interpretable \cite{guu2020retrieval,louis2024interpretable}, and adaptable \cite{xu2023retrieval,zakka2024almanac}. 
Traditional practices \cite{karpukhin2020dense,min2019knowledge} often involve providing top-retrieved passages as the input context to LLMs without discrimination. 
However, imperfect retrieval systems frequently yield irrelevant content. Furthermore, indiscriminately feeding all retrieved content to LLMs will cause input redundancy, imposing a high computational cost and making them prone to hallucination \cite{shi2023large}.

Ideally, LLMs should be grounded on supporting content that is both highly helpful to address user input and sufficiently concise to facilitate inference speed. 
However, it is practically impossible for imperfect retrieval systems to achieve such an ideal grounding solely \cite{wang2023learning}. 
In fact, top-retrieved passages usually compose supporting and distracting content, inflicting a heavy blow on LLMs trained with high-quality corpora to generate the correct output. 
This motivates us to develop an evidence extractor, that aims at extracting supporting content while filtering out\, distracting\, content. 

Recently, a pioneering study, \textsc{FilCo} \cite{wang2023learning}, attempts to retrieve chunking document content with sentence precision via three filters, \textit{i.e.,} StrInc, Lexical, and CXMI. 
Then, it trains a context filtering model, using context filtered by the above three measures as ground truth. 
Despite effectiveness, current context-filtering methods have several limitations: 
\textbf{(1) Hand-crafted Context Filtering.} Manually designed context-filtering measures typically require domain knowledge, which can hardly be adaptable to diverse downstream tasks with limited supervision. 
\textbf{(2) Disruptive Chunking on Context.} The use of chunking strategies may be ineffective as rule-based splitting on context usually cannot preserve its original semantics and often produces semantically deficient text blocks. 
\textbf{(3) Skewed Distribution in Length.} The length of supporting content in top-retrieved passages may vary largely across different samples. Hence, learning to filter context sentence-wise is biased\, toward\, skewed\, length\, distribution.

Given these limitations, an interesting question arises: \textit{Now that heuristic-based augmentation\footnote{Previous methods generally create training signals based on heuristics. We denoted it as heuristic-based augmentation.} suffers from several issues, can we develop a model-based augmentation method free of the above problems?}
Inspired by the recent success of self-alignment \cite{li2023self,zhang2024self,liang2024learning}, self-aligned learning utilizes the model to improve itself and aligns its response with desired properties, 
which can mitigate the heavy reliance on hand-crafted context filtering, rule-based context chunking, and sentence-wise filter learning. 

\begin{figure}[t]
    \centering
    \includegraphics[width=1.\linewidth]{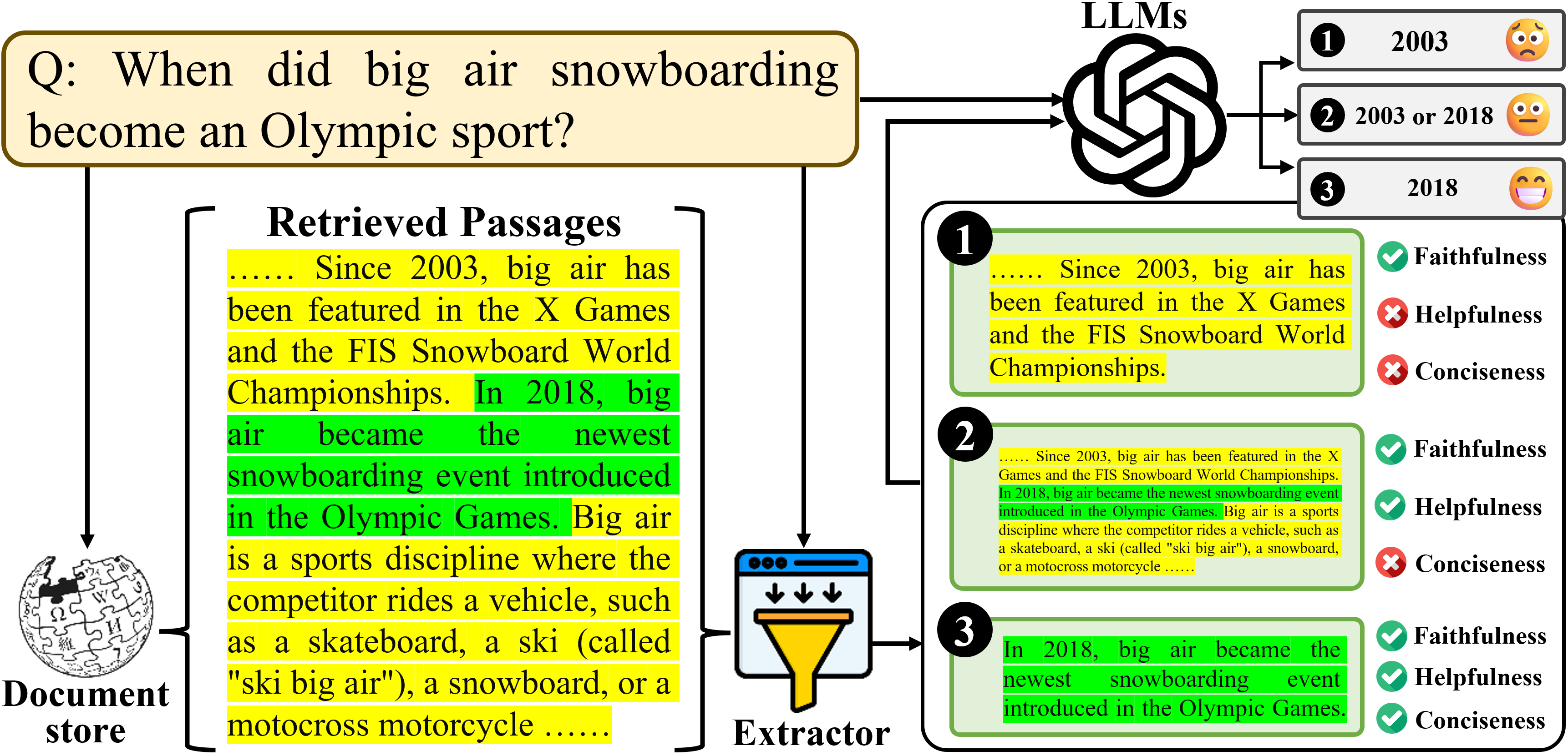}
    \caption{The RAG pipeline with the evidence extractor, in which the supporting content and the distracting content are marked in \textbf{\textcolor[RGB]{0,160,0}{green}}\, and\, \textbf{\textcolor[RGB]{194,144,0}{yellow}},\, respectively.} 
    \label{fig:pipline}
\end{figure}
Given the extracted evidence, a question arises again: \textit{How to evaluate the quality of evidence properly?}
In principle, the evidence should be faithful (\textit{i.e.,} avoiding intrinsic hallucination) to the retrieved passages \cite{rashkin2021increasing,maynez2020faithfulness}, helpful in addressing the user input \cite{adlakha2023evaluating}, and concise to facilitate the inference speed \cite{ko2024evidence}. Figure \ref{fig:pipline} shows three representative scenarios: 
\textbf{(1)} When the evidence only favors faithfulness, LLMs may generate an incorrect answer; 
\textbf{(2)} When the evidence further favors helpfulness but lacks conciseness, LLMs' attention may be distracted by noise; 
\textbf{(3)} When the evidence favors all three criteria, LLMs can generate confidently with low computational costs.

In this paper, we propose a model-based evidence extraction learning framework, \textbf{\textsc{SEER}}, \textbf{S}elf-Aligned \textbf{E}vidence \textbf{E}xtraction for \textbf{R}etrieval-Augmented Generation.
Specifically, it consists of three primary stages: 
\textbf{(1) Evidence Extraction:} To mitigate the issues above, we propose extracting diversified evidence with semantic consistency and varying length through response sampling, offering sufficient preference data for alignment.
\textbf{(2) Expert Assessment:} For each extracted evidence, we construct a quadruple, QuadQARE, made up of query, answer, passage, and evidence. Then, we devise three experts to assess the quality of each extracted evidence \textit{w.r.t.} three primary criteria.
Given these scores, we propose smoothing CoV-Weighting, which explicitly leverages the statistics to estimate their relative weighting and result in the CoV-Weighted scores. 
\textbf{(3) Self-Alignment:} With a ranking list of extracted evidence and their smoothing CoV-weighted scores, a question remains:  \textit{How to optimize extraction preference with the ranking position?}
To this end, we propose a listwise-aware Lambda Preference Optimization method, LPO, assigning each preference pair with a listwise-aware weight scaled by the gain in Reciprocal Rank from swapping the position of two evidence \cite{donmez2009local,burges2006learning,wang2018lambdaloss}.

It is worth mentioning that \textbf{SEER} is a \textbf{criterion-agnostic} framework and can employ any off-the-shelf expert. 
In this work, we use faithfulness, helpfulness, and conciseness, which are regarded as three primary criteria for assessing the quality of evidence \cite{maynez2020faithfulness,rashkin2021increasing,adlakha2023evaluating,ko2024evidence}.
Our \textbf{main contributions} can be\, summarized\, as\, four-folds: 
\begin{itemize}[leftmargin=*]
    \item We propose a novel evidence extraction learning framework, \textbf{SEER}, which leverages preference data augmented by the model to improve performance and also is free of the arduous workforce. 
    \item We devise three experts to assess the quality of the evidence and design a smoothing CoV-weighting schema to get the overall assessment, meeting the property of being criterion-agnostic.
    \item We propose a listwise-aware preference optimization method, LPO, which seamlessly brings the ranking position signals into preference learning. 
    \item Extensive experiments on three benchmark datasets show that our method can considerably improve QA performance, enhance the quality of evidence, as well as reduce computational costs.
\end{itemize}

\section{Preliminaries}
\label{sec:prelimi}
\subsection{Problem Formulation}
In this task, we are given a base extractor $\mathcal{E}$, and a fixed generator $\mathcal{G}$, where we choose Llama2-7b-Chat \cite{DBLP:journals/corr/abs-2307-09288} as the backbone for the base extractor $\mathcal{E}$. 
For a given query $q$ and its corresponding golden answer $a$, we assume a set of retrieved passages $P=\{p_i\}_{i=1}^K$, where $K$ is the retrieved size. 
Here, we aim to fine-tune the base extractor $\mathcal{E}$ via self-alignment to get the aligned extractor $\tilde{\mathcal{E}}$, for the generator $\mathcal{G}$ to leverage the better evidence and achieve superior performance:
\begin{equation}
    e \sim \tilde{\mathcal{E}}(\cdot|q\oplus P), \quad o \sim \mathcal{G}(\cdot|q\oplus e),
\end{equation}
where $e$ and $o$ denote the extracted evidence and the generated output, respectively; $\oplus$ denotes the concatenation operation; $q$ is the given user query.

\subsection{Augmentation Analysis}
\label{sec:analysisAug}
\begin{figure}[t]
    \centering
    \includegraphics[width=1.\linewidth]{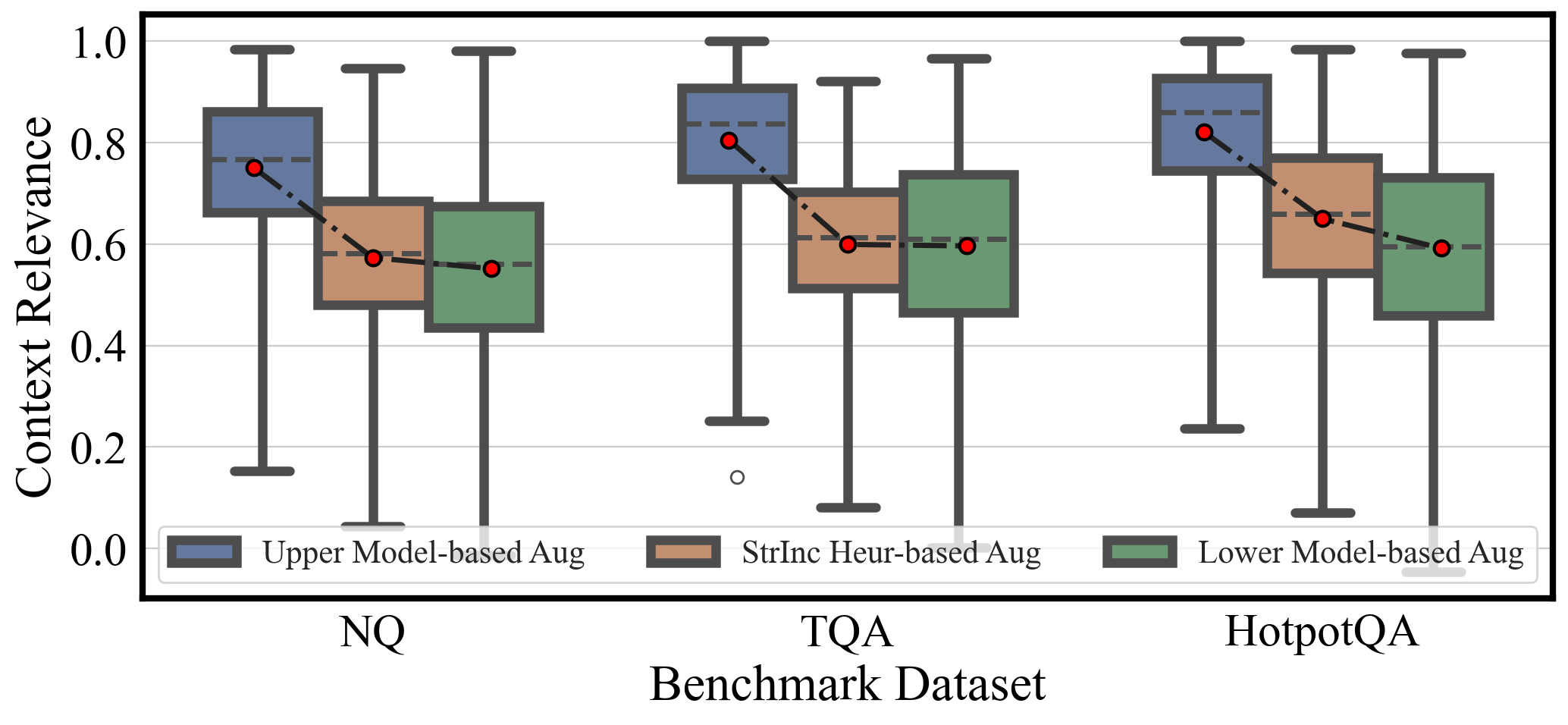}
    \caption{Comparison between model-based and heuristic-based augmentation \textit{w.r.t.}  context relevance.} 
    \label{fig:analysisAug}
\end{figure}

As stated in Section \ref{sec:intro}, heuristic-based augmentation suffers from several issues, 
which severely hinders the optimization of context filtering. 
To verify the above claim, we compare the context relevance between heuristic-based and model-based augmentation, 
where the \textbf{context relevance} is the cosine similarity between the extracted evidence and the user query\footnote{We employ the SBERT-NLI-base model \citet{reimers-2019-sentence-bert} (denoted as SBERT) to encode the extracted evidence and the user query into sentence embedding vectors.}. 
Here, we use StrInc as the representative heuristic-based augmentation method (abbreviated as ``\textbf{StrInc Heur-based Aug}''), as it usually performs best on QA tasks according to \cite{wang2023learning}. 
On the other hand, we perform model-based augmentation by response sampling (More details can be seen in \S\ref{sec:evidence}). 
We take the best-performing extracted evidence for each QA pair as ``\textbf{Upper Model-based Aug}'' while the worst-performing one as ``\textbf{Lower Model-based Aug}''. 
We experiment on three datasets, \textit{i.e.,} NQ, TQA, and HotpotQA.
Figure \ref{fig:analysisAug} shows that: \textbf{(1)} The context relevance of Upper Model-based Aug is consistently higher than that of StrInc Heur-based Aug. 
\textbf{(2)} The context relevance of {StrInc Heur-based Aug} generally lies in the middle of Upper and Lower Model-based Aug. 
From the above observations, our claim is well-validated, as model-based augmentation shows a larger potential than heuristic-based one. 
Therefore, it is valuable to conduct model-based augmentation for better performance.

\section{Methodology}
\label{sec:method}
\begin{figure*}[t]
    \centering
    \includegraphics[width=1.0\linewidth]{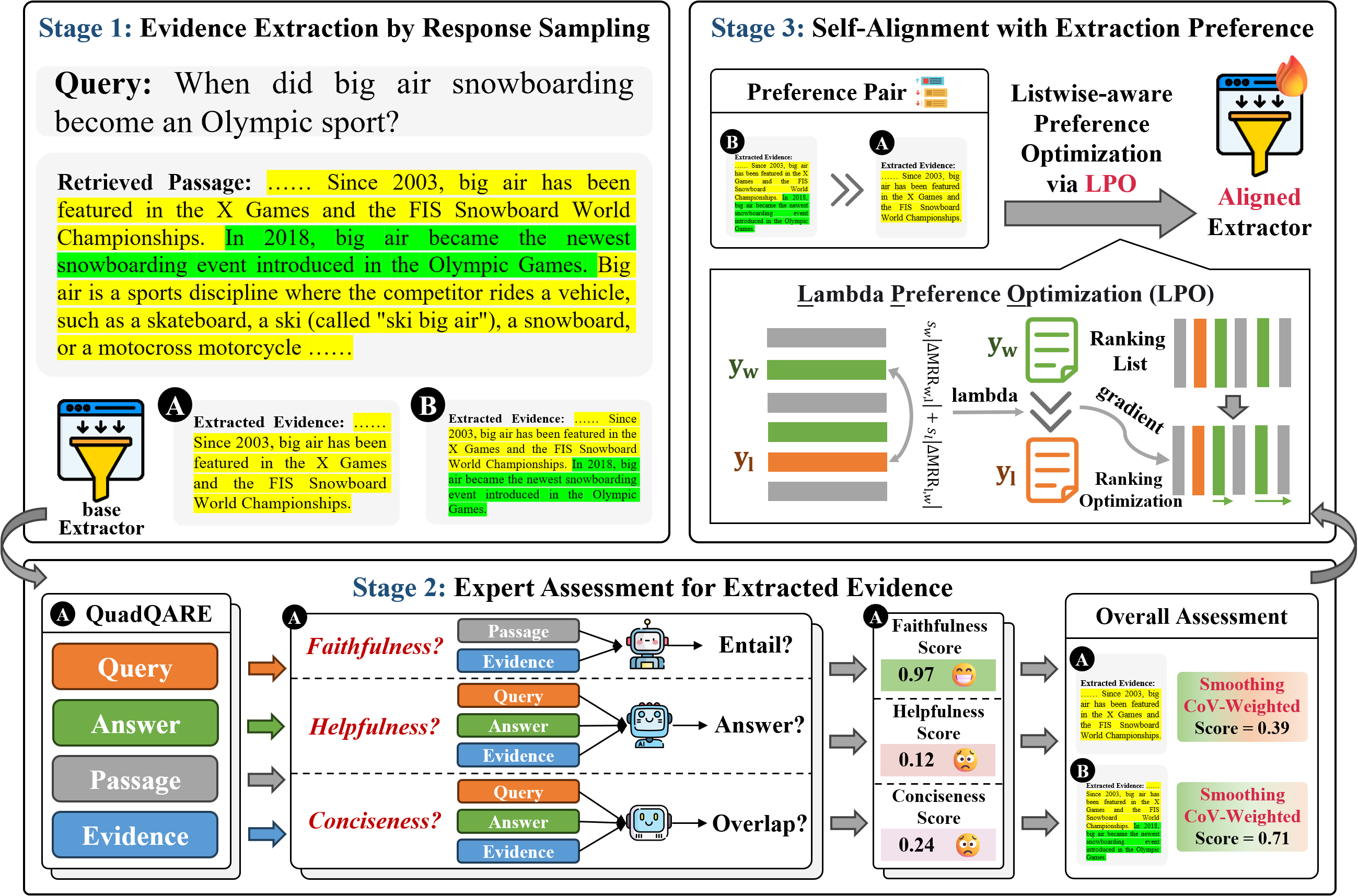}
    \caption{The overall system framework of our \textbf{SEER}, which mainly consists of three modeling stages.} 
    \label{fig:framework}
\end{figure*}

Figure \ref{fig:framework} depicts the overall framework of \textbf{SEER}, composing three key stages: \textbf{(1) Evidence Extraction} (\S\ref{sec:evidence}), which extracts evidence via response sampling. 
\textbf{(2) Expert Assessment} (\S\ref{sec:expert}), which assesses the quality of evidence.
\textbf{(3) Self-Alignment} (\S\ref{sec:self_alignment}), which aligns the extractor with extraction preference. 
The learning algorithm of our proposed method can be seen in Appendix \ref{appendix:seer_alg} in Algorithm \ref{alg:seer_alg}.

\subsection{Evidence Extraction Stage}
\label{sec:evidence}
As stated in Section \ref{sec:intro}, heuristic-based augmentation \cite{wang2023learning} suffers from several issues. 
An empirical study (\S\ref{sec:analysisAug}) further indicates that model-based augmentation is more beneficial for performance improvement than heuristic-based augmentation.
Hence, we aim to utilize the base extractor $\mathcal{E}$ to improve itself and align it with desired properties. 
To this end, we probe into its evidence extraction preference by response sampling for preference data collection. 
Specifically, given a query $q$ and its retrieved passage ${P}$, we prompt the model to generate multiple candidate extracted evidence $\{e_i\}_{i=1}^M$ via response sampling $e_* \sim {\mathcal{E}}(\cdot|q\oplus P)$, where $M$ is the sample size. 

However, LLMs often tend to be overconfident in their knowledge \cite{xiong2023can}. 
As such, the response distribution typically follows a power-law, where head responses occupy a large portion of extracted evidence while long-tail ones are very sparse. 
Directly using the power-law response distribution for alignment would cause preference optimization to be biased toward head responses. 
Hence, we remove duplicates and obtain the uniformly distributed set, \textit{i.e.,} $\{e_i\}_{i=1}^N$, where we use n-gram similarity \cite{DBLP:conf/spire/Kondrak05} to detect duplicates and $N$ is the remaining size.
In practice, we find using the uniform response distribution does matter for alignment to reach higher performance.
\subsection{Expert Assessment Stage}
\label{sec:expert}
Although the base extractor is able to extract evidence, its output might be unfaithful, unhelpful, as well as unconcise, which are regarded as three primary factors that hinder the quality of evidence \cite{maynez2020faithfulness,rashkin2021increasing,adlakha2023evaluating,ko2024evidence}. 
Considering the above issues, we devise \textbf{three experts} to assess the quality of extracted evidence \textit{w.r.t.} faithfulness, helpfulness, and conciseness\footnote{We use the term ``\textbf{{oracle}}'' to denote three primary criteria.}, respectively. 
Subsequently, given multiple scores for each extracted evidence, we devise a \textbf{smoothing CoV-Weighting} schema in order to get the overall assessment score. 

\nosection{Obtaining Oracle Scores}
For expert assessment stage, we first collect a set of QuadQARE $<q, a, P, e>$, where a \underline{Quad}ruple is composed of \underline{Q}uery $q$, \underline{A}nswer $a$, \underline{R}etrieved passage $P$, and extracted \underline{E}vidence $e$. 
Afterwards, we design three plug-and-play experts to parallelly assess the quality of extracted evidence,\, from\, different\, aspects:
\begin{itemize}[leftmargin=*]
    \item \textbf{Faithfulness Expert.} It focuses on the faithfulness of each extracted evidence. Toward this end, we adopt an advanced NLI model,  
    \textsc{AlignScore}\footnote{We use \textsc{AlignScore}-large for faithfulness assessment.} \cite{DBLP:conf/acl/ZhaYLH23}, 
    to evaluate the consistency between the retrieved passage $P$ and extracted evidence $e$ in terms of hallucination.
    Specifically, we treat the retrieved passage and the corresponding extracted evidence as the premise and hypothesis, respectively. 
    Then, we employ \textsc{AlignScore} to measure to what extent the extracted evidence $e$ could be entailed by the retrieved passage $P$, which can be formulated as:
    \begin{equation} 
     s^f = \textsc{AlignScore}(P, e), \label{eq:faith_score}
    \end{equation}
    where $s^f \in [0, 1]$  is the faithfulness score. If the hypothesis $e$ is faithful to the premise $P$, then \quad \quad the score is close to 1, otherwise, it is close to 0.

    \item \textbf{Helpfulness Expert.} It examines the helpfulness of each extracted evidence candidate in terms of output improvement. 
    In other words, it checks whether the extracted evidence $e$ contributes to the model's output improvement when utilized as input. 
    Specifically, we assess its potential influence on LLMs by calculating the change in the log probability of generating the golden answer $a$ between the model's output before and after\, the\, inclusion\, of\, the\, extracted\, evidence $e$:
    \begin{equation}
     s^h = \mathrm{Sig}\left(\log\frac{\prod f(a|q\oplus e)}{\prod f(a|q)}\right),  \label{eq:help_score}
    \end{equation}
    where $s^h \in [0, 1]$  is the helpfulness score, $f(\cdot)$ is the helpfulness expert\footnote{We employ Flan-T5-XL for helpfulness assessment.}, 
    $\mathrm{Sig}(\cdot)$ is the sigmoid function. Similarly, if the extracted evidence $e$ is helpful for LLMs to output the golden answer $a$, the score is close to 1, otherwise, it is close to 0.

    \item \textbf{Conciseness Expert.} If only the above two experts are considered, the aligned extractor can easily be achieved by directly treating the retrieved passage as evidence.
    To avoid such a trivial solution, we further measure the conciseness of the extracted evidence $e$. 
    Towards this end, we first convert the query $q$ and the golden answer $a$ into the full-length answer\footnote{The full-length answer is generated by transforming the question and its corresponding answer into a declarative statement \cite{pal2019answering,jain2021natural}.} $t$, which represents minimal information for the need to answer the query. 
    Subsequently, we leverage SBERT \cite{reimers-2019-sentence-bert} to measure to what extent the semantic  overlap between the full-length answer and the extracted evidence:
    \begin{equation}
     s^c = \mathrm{SBERT}_{\mathrm{cosine}}(t,e), \label{eq:concise_score}
    \end{equation}
    where $s^c \in [-1, 1]$  is the conciseness score that is measured by cosine similarity between the sentence embedding of $t$ and $e$, $t$ is a full-length answer. 
    In this work, we prompt GPT-3.5-turbo to generate a full-length answer $t$ given the query $q$ and its answer $a$. More details about full-length answer generation can be seen in Appendix \ref{appendix:full}.
\end{itemize}

\vspace{-0.2cm}
\nosection{Weighting Oracle Scores} Having obtained the oracle scores, a question naturally arises: \textit{How to get the overall assessment for each extracted evidence?} 
A straightforward way is to compute the average of the oracle scores. However, equal weighting might not result in optimal alignment, since the learning difficulty is inconsistent. 
Therefore, the weights should match the learning difficulty to guide the preference optimization process. 
Given this, we propose smoothing CoV-weighting, leveraging the variability\, of\, the\, scores\, in\, relation\, to\, the\, mean:
\begin{equation}
     c^f = {\sigma^f}/{\mu^f}, \label{eq:sigma_mu} 
\end{equation}
where $\sigma^f$ and $\mu^f$ denote the standard deviation and the mean of faithfulness score $s^f$, $c^f$ is the \underline{C}oefficient \underline{o}f \underline{V}ariation (CoV) whose value is independent of the magnitude. 
As such, CoV can decouple the score magnitude from the score weighting, so a type of score with a small magnitude may still be relatively impactful when it is variant \cite{groenendijk2021multi}. 
Analogously, we obtain the CoV of the helpfulness and conciseness score, \textit{i.e.,} $c^h$ and $c^c$. 
Moreover, we employ the softmax function with temperature on the coefficient of variation of these scores, which controls the smoothness of the score weight to avoid\, abnormal\, score\, weight:
\begin{equation}
    \alpha^f = \frac{\mathrm{exp}({{c^f}/{\tau}})}{\sum_* \mathrm{exp}({{c^*}/{\tau}})},
\end{equation}
where $\alpha^f$ is the faithfulness score weight, $\tau$ is the temperature. 
Analogously, we obtain the helpfulness and conciseness score weight, \textit{i.e.,} $\alpha^h$ and $\alpha^c$. Then, the\, CoV-weighted\, score\, can\, be defined as:
\begin{equation}
    s = \alpha^f s^f + \alpha^h s^h + \alpha^c s^c,  \label{eq:cov_score} 
\end{equation}
where the score weight increases when the std increases or the mean decreases,
ensuring more optimization proceeds when the score is more variant.
\subsection{Self-Alignment Stage} 
\label{sec:self_alignment}
After obtaining the preference data over all candidates $\mathcal{D}=\{(q\oplus P, e_i, e_j)| 1\leq i,j \leq N, s_i>s_j\}$, 
where each tuple represents a choice preference between winning and losing extracted evidence, we proceed to the stage of alignment tuning for improving faithfulness, helpfulness, and conciseness. 
For alignment training, previous works commonly adopt Proximal Policy Optimization (PPO) \cite{schulman2017proximal} or Direct Preference Optimization (DPO) \cite{rafailov2024direct}. 
However, PPO cannot perceive the ranking position and DPO treats all preference pairs indiscriminately. Due to the above drawbacks, both of them cannot result in optimal alignment.
Inspired by the Lambdaloss method \cite{donmez2009local,burges2006learning,wang2018lambdaloss}, we propose a listwise-aware Lambda Preference Optimization algorithm, LPO, which seamlessly brings the ranking position into DPO\, by\, assigning\, a\, lambda\, weight\, to each pair:
\begin{normalsize}
    \begin{equation}
        \begin{aligned}
            &\mathcal{L}(\pi_\theta;\pi_{\mathrm{ref}},\lambda_{w,l})_{\mathrm{LPO}}  =  -\mathbb{E}_{(x, y_w, y_l) \sim \mathcal{D}}  \\ &  \,\, \left[ \lambda_{w,l} \log \mathrm{Sig}\left(\beta \frac{\pi_\theta(y_w|x)}{\pi_{\mathrm{ref}}(y_w|x)} 
                -\beta\frac{\pi_\theta(y_l|x)}{\pi_{\mathrm{ref}}(y_l|x)}\right) \right],
        \end{aligned} \label{eq:lpo_loss}
    \end{equation}
\end{normalsize}where $\pi_{\theta}=\tilde{\mathcal{E}}, \pi_{\mathrm{ref}}={\mathcal{E}}, x=q\oplus P, y_w,y_l=e_i,e_j$. 
We implement the lambda weight $\lambda_{w,l}$ for Mean Reciprocal Rank (MRR), \textit{i.e.,} measuring the gain in Reciprocal Rank from swapping the position of two candidates, which can be formulated as follows:
\begin{equation}
    \lambda_{w,l} = s_w\Delta\mathrm{MRR}_{w,l}+s_l\Delta\mathrm{MRR}_{l,w}, \label{eq:lambda_weight}
\end{equation}
where $\Delta\mathrm{MRR}_{w,l}=\frac{1}{r_w}-\frac{1}{r_l}$, $r_w$ is the rank position of $y_w$ in the ranking permutation induced by the smoothing CoV-weighted score $s$. 
Thus, by introducing the lambda weight, LPO becomes a listwise-aware method. LPO is designed to work with any ranking metric, as long as the lambda weight can be defined, \textit{e.g.,} NDCG \cite{DBLP:journals/corr/abs-2402-01878}. 
Here, we implement LPO to optimize a well-founded ranking metric MRR because it is simple\, yet\, effective.

\begin{table*}[t]
  \renewcommand\arraystretch{1.5}
  \tabcolsep=0.112cm
  \fontsize{10}{10}\selectfont
  \begin{tabular}{cccc|cc|cccc}
    \toprule
    \multicolumn{1}{c}{\multirow{2}{*}{\centering \bf Datasets}} 
    & \multicolumn{1}{c}{\multirow{2}{*}{\centering \bf Generators}} 
    & \multicolumn{1}{c}{\multirow{2}{*}{\centering \bf Metrics}} 
    & \multicolumn{1}{c|}{\centering \bf WE}
    & \multicolumn{2}{c|}{\centering \bf CGE} 
    & \multicolumn{4}{c}{\centering \bf FGE} \\
    \cline{4-10}
    & & & \multirow{1}{*}[-0.5ex]{\centering Zero} &  \multirow{1}{*}[-0.5ex]{\centering Full}  &  \multirow{1}{*}[-0.5ex]{\centering SeleCtx} &  \multirow{1}{*}[-0.5ex]{\centering LLM-Embedder} &  \multirow{1}{*}[-0.5ex]{\centering Bge-Reranker} & \multirow{1}{*}[-0.5ex]{\centering \textsc{FilCo}} & \multirow{1}{*}[-0.5ex]{\centering \textbf{SEER}} \\ 
    \midrule

     &  \multirow{2}{*} {\centering Flan-T5} & EM & 0.0934 & \underline{0.4137} & 0.2853 & 0.3953  & 0.4089 & 0.3809 & \textbf{0.4322}\\
    &   & Tok  & 0 & 732 & 290 & 147 & 148 & \textbf{62} &  \underline{89}\\
    \rowcolor{gray!12}
     \cellcolor{white!20} &  &  EM & 0.2695 & \underline{0.4382} & 0.3850 & 0.4208 &  0.4202 & 0.4061 & \textbf{0.4549} \\
    \rowcolor{gray!12}
   \cellcolor{white!20} \multirow{-4}{*} {\centering \textbf{NQ}} &  \multirow{-2}{*} {\centering  Llama2} &  Tok & 0 & 804 & 319 & 160 & 162 & \textbf{67} &  \underline{95}\\
    \hline

     &  \multirow{2}{*} {\centering Flan-T5} & EM  & 0.2621 & 0.6320 & 0.5022 &  0.5689 & 0.6227 &  \underline{0.6431} & \textbf{0.6503} \\
    &   & Tok  & 0 & 760 & 306 & {152}  & 153 &  \underline{130} & \textbf{121}\\
    \rowcolor{gray!12}
    \cellcolor{white!20} &   & EM & 0.4898 & 0.6571 & 0.6061 & 0.6239 & 0.6581 & \underline{0.6599} & \textbf{0.6711} \\
    \rowcolor{gray!12}
    \cellcolor{white!20} \multirow{-4}{*} {\centering \textbf{TQA}} &  \multirow{-2}{*} {\centering  Llama2} & Tok & 0 & 813 & 331 & {161} & 163 &  \underline{137} & \textbf{133}\\
    \hline

     &  \multirow{2}{*} {\centering Flan-T5} & $\mathrm{F_1}$  & 0.5289 & \textbf{0.5702} & 0.5127 &  0.5532 & 0.5608 &  0.5535 & \underline{0.5615}\\
    &   & Tok & 0 & 765 & 313 &  154 & 153 & \textbf{56} &  \underline{58}\\
    \rowcolor{gray!12}
    \cellcolor{white!20} &   & $\mathrm{F_1}$ & 0.5471 & 0.5826 & 0.5328 & 0.5703 & 0.5734 &  \underline{0.5977} & \textbf{0.6040}\\
    \rowcolor{gray!12}
   \cellcolor{white!20} \multirow{-4}{*} {\centering \textbf{HotpotQA}} & \multirow{-2}{*} {\centering  Llama2}  & Tok  & 0 & 821 & 337 & 165 & 164 & \textbf{59} & \underline{62}\\

    \bottomrule
  \end{tabular}
  \caption{QA performance comparison, where the best results are \textbf{boldfaced} and the second-best results are \underline{underlined}, in each row. `Tok' is the average length of extracted evidence fed into generators, where the smaller the value, the lower the computational cost. All improvements are significant with $p$-value < 0.01 according\, to\, $t$-test.}
  \label{tab:overall}
\end{table*}
\section{Experiments}
\label{sec:experiment}
In this section, we conduct extensive experiments on three QA benchmark datasets to answer the following Research Questions (\textbf{RQs}): 
\textbf{RQ1:} How does our model contribute to QA accuracy compared with other state-of-the-art methods? 
\textbf{RQ2:} Can LPO facilitate the generation of more faithful, helpful, and concise evidence? 
\textbf{RQ3:} Can our model perform robustly to noise from irrelevant passages?
\textbf{RQ4:} How effective are the key settings in our model,\, such as smoothing CoV-weighting?
\subsection{Experimental Settings}

\nosection{Datasets and Metrics} We experiment on three benchmark QA datasets, NaturalQuestions (NQ) \cite{kwiatkowski2019natural}, TriviaQA (TQA) \cite{joshi2017triviaqa}, and HotpotQA \cite{yang2018hotpotqa}. 
Following \citet{wang2023learning}, we use the processed version \cite{lee2019latent} of NQ for experiments, discarding answers with more than 5 tokens.
As NQ and TQA belong to the extractive QA task, we use Exact Match (EM) as their evaluation metric, where a score of 1 is assigned if at least one among multiple correct answers appears in the response of the QA model; otherwise, the score is 0. 
While HotpotQA belongs to the abstractive QA task, we employ unigram $\mathrm{F}_1$ to evaluate answer correctness. 
As the test set for HotpotQA is unavailable, we report the dev set results. The detailed statistics of datasets are summarized in Appendix \ref{appendix:implementation} in Table \ref{tab:dataset}.

\nosection{Baseline Methods} There are three types of baselines: 
\textbf{(1) Without Evidence (WE)} includes \textbf{(i) Zero-shot (Zero)} that does not pass any evidence to LLMs. 
\textbf{(2) Coarse-grained Evidence ({CGE})} includes \textbf{(i) Full Passage (Full)} that directly passes the top-retrieved passage to LLMs, \textbf{(ii) Select-Context (SeleCtx)} \cite{li2023compressing} that identifies and prunes redundancy in the top-retrieved passage based on perplexity. 
\textbf{(3) Fine-grained Evidence (FGE)} includes \textbf{(i) LLM-Embedder} \cite{zhang2023retrieve} that extracts the sub-passages with the highest similarity to the query from the top-retrieved passage, 
\textbf{(ii) Bge-Reranker-Large (Bge-Reranker)} \cite{xiao2023c} that reorders all sub-passages in the top-retrieved passage and uses the top-ranked sentence as evidence,  
\textbf{(iii) \textsc{FilCo}} \cite{wang2023learning} that learns to filter the retrieved passage with sentence precision leveraging heuristic-based augmentation to label ground-truth.

\nosection{Generators for QA} 
To measure the efficacy of the evidence extracted by $\textbf{SEER}$ and other competitive baselines,
we employ two different generators, \textit{i.e.,}, \textbf{Flan-T5-XL} \cite{chung2024scaling} and \textbf{Llama2-7B-Chat} \cite{DBLP:journals/corr/abs-2307-09288}, for QA evaluation\footnote{In what follows, we use Flan-T5 and Llama2 to represent Flan-T5-XL and Llama2-7B-Chat, respectively, for brevity.}.

\nosection{Implementation Details} Following \citet{wang2023learning}, we use the adversarial Dense Passage Retriever (DPR) \cite{karpukhin2020dense} to retrieve the top-5 passages from all Wikipedia passages. 
For each <user query $q$, retrieved passage $P$> pair, we set the sample size $M$ as 10.
For the temperature coefficient of smoothing CoV-weighting, we tune it within the range of $\{0.2,0.5,1.0,2.0,5.0\}$.
We employ Llama2-7B-Chat \cite{DBLP:journals/corr/abs-2307-09288} as the base extractor $\mathcal{E}$ and fine-tune it on the constructed preference data for 2 epochs to get the aligned extractor $\tilde{\mathcal{E}}$. 
We adopt greedy decoding for evidence extraction and output generation. More implementation details are shown\, in\, Appendix \ref{appendix:implementation}. 
\begin{figure}[t]
    \centering  
    \subfigure[NQ dataset.]{
        \includegraphics[width=1.0\linewidth]{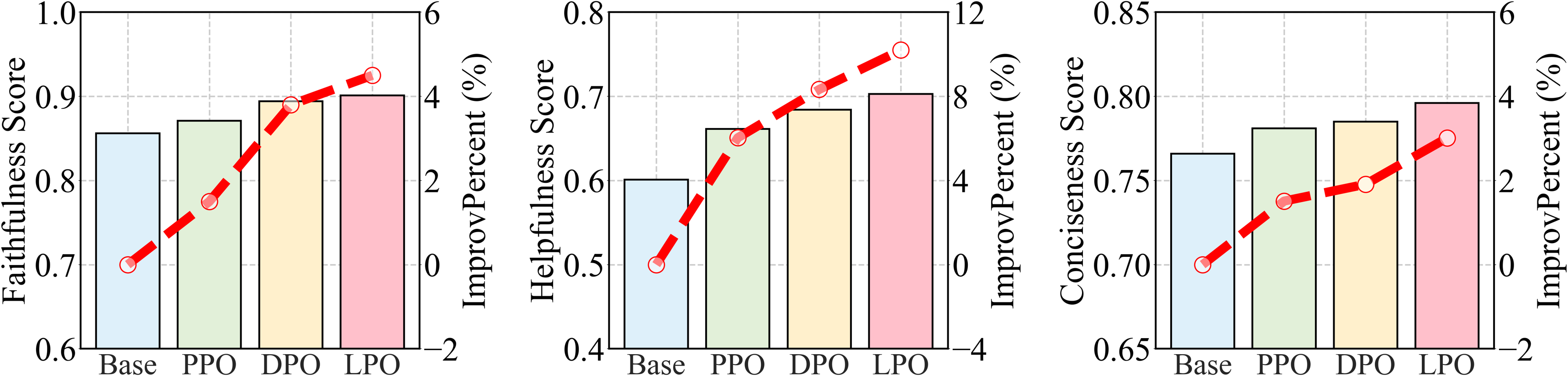}\label{fig:NQ_align}}
    \subfigure[TQA dataset.]{
        \includegraphics[width=1.0\linewidth]{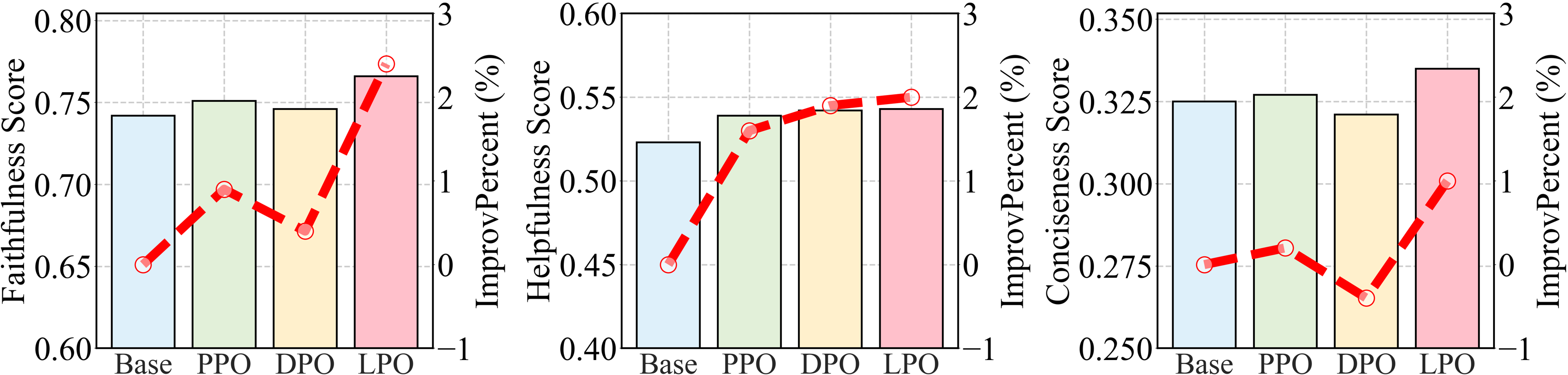}\label{fig:TQA_align}}
    \subfigure[HotpotQA dataset.]{
        \includegraphics[width=1.0\linewidth]{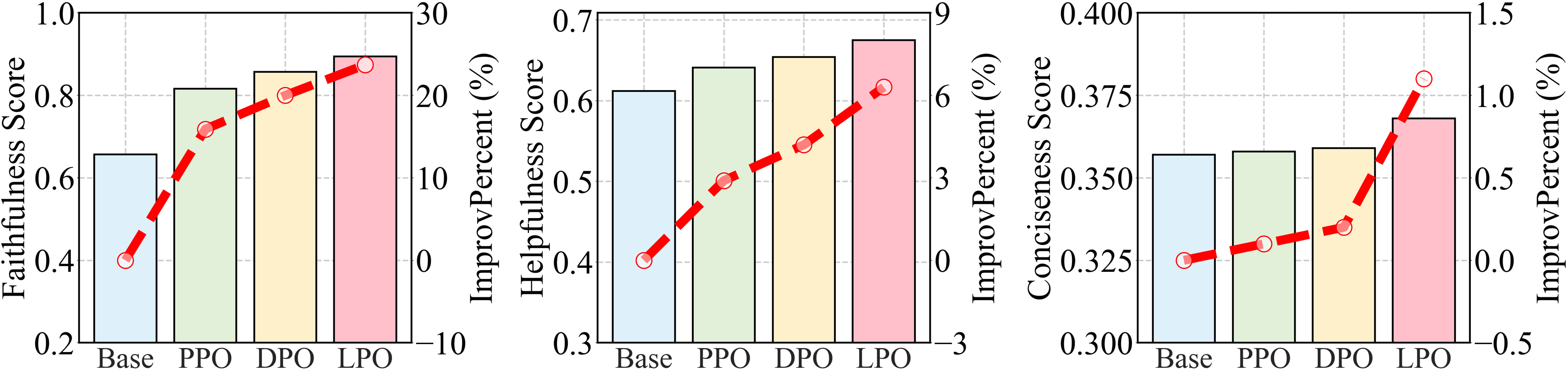}\label{fig:Hot_align}}
    \caption{Alignment performance \textit{w.r.t.} faithfulness, helpfulness, and conciseness. The bar represents the oracle scores, while the line denotes the percentage of performance improvement in comparison with the Base.}
    \label{fig:align_study}
\end{figure}
\subsection{Model Comparison (RQ1)}
To examine the impact of evidence extraction on the final RAG performance, we experimented on three benchmark QA datasets, where we prepended the extracted evidence before the user query and then input it together into the generator. 
Besides, we use the tokenizer of Flan-T5 and Llama2 to convert the extracted evidence into a list of subwords and then calculate the length of the list, where the length is adopted as a metric (denoted by `\textbf{Tok}') measuring the computational burden to a large extend. 
Table \ref{tab:overall} shows the final RAG performance of different baseline evidence extraction methods and our proposed \textbf{SEER}. From the experimental results, we mainly have the following observations:
\begin{itemize}[leftmargin=*]
    \item In all cases, \textbf{SEER} outperforms \textsc{FilCo} by a large margin, indicating the superiority of model-based augmentation that can provide more informative signals than heuristic-based augmentation. 
    For example, \textbf{SEER} achieves 13.5\% and 12.0\% improvements over \textsc{FilCo} in the NQ dataset with Flan-T5 and Llama2 generators, respectively, while the average evidence\, length is very\, close.
    \item Optimizing the three primary criteria for evidence extraction (\textit{i.e.,} faithfulness, helpfulness, and conciseness) yields such impressive performance improvements, considering most baselines come from studies in recent two years. 
    This demonstrates that these three properties strongly agree with the evidence quality in RAG, while current methods might not satisfy all of them simultaneously,\, which leads to\, inferior\, results.
    \item Comparing different baselines, it is not surprising the method without evidence performs the worst. Secondly, methods with fine-grained evidence do not always perform better than ones with coarse-grained evidence. 
    Specifically, the `\textbf{Full}' method generally performs well, as it preserves retrieved passages complete, while some FGE methods (\textit{e.g.,} LLM-Embedder and Bge-Ranker) might lose key information in the process of evidence extraction, but it takes much more time for generation due to the long context.
    Last but not least, our \textbf{SEER} considerably outperforms the `\textbf{Full}' method in most cases, where the average improvement on the three datasets is 2.58\% \textit{w.r.t.} QA accuracy, but the average length of evidence fed into generators is reduced by a factor of 9.25.
\end{itemize}

\begin{figure}[t]
    \centering  
    \subfigure[NQ dataset.]{
        \includegraphics[width=1.0\linewidth]{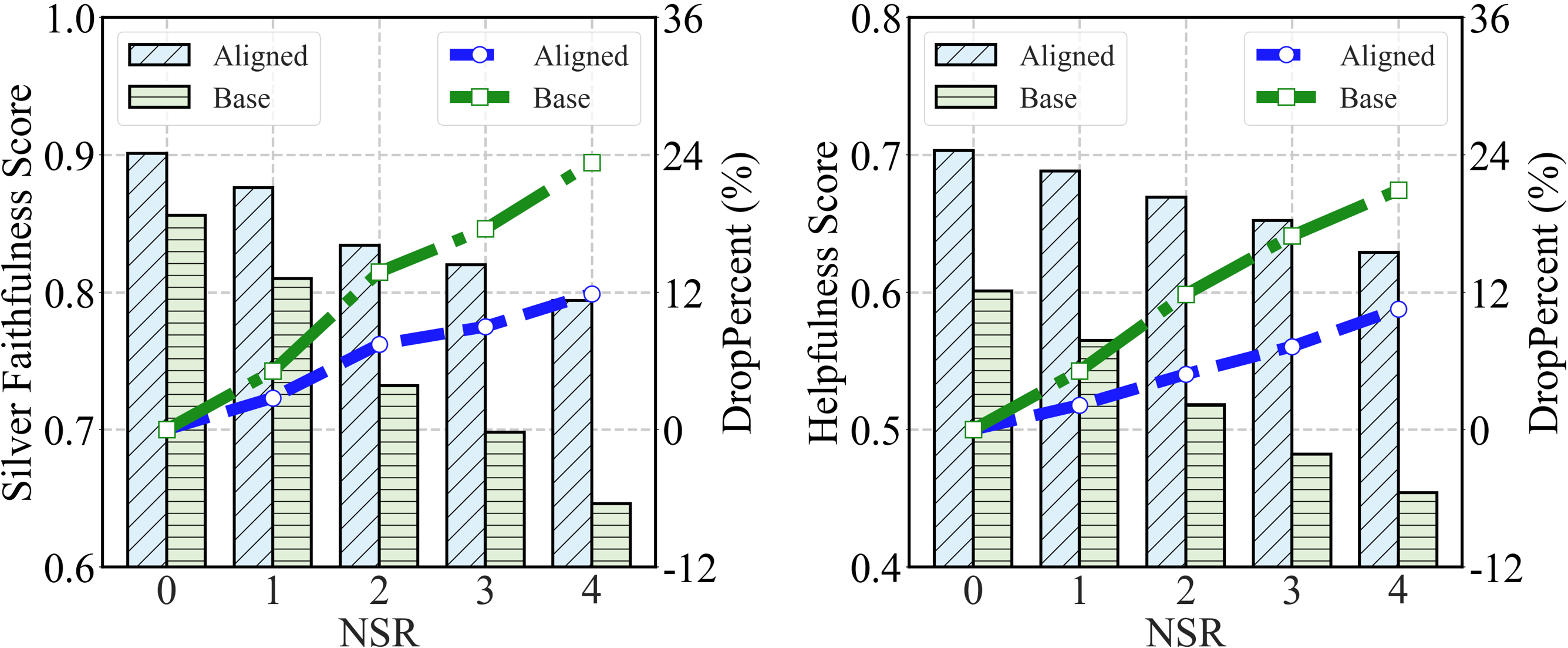}\label{fig:NQ_robust}}
    \subfigure[TQA dataset.]{
        \includegraphics[width=1.0\linewidth]{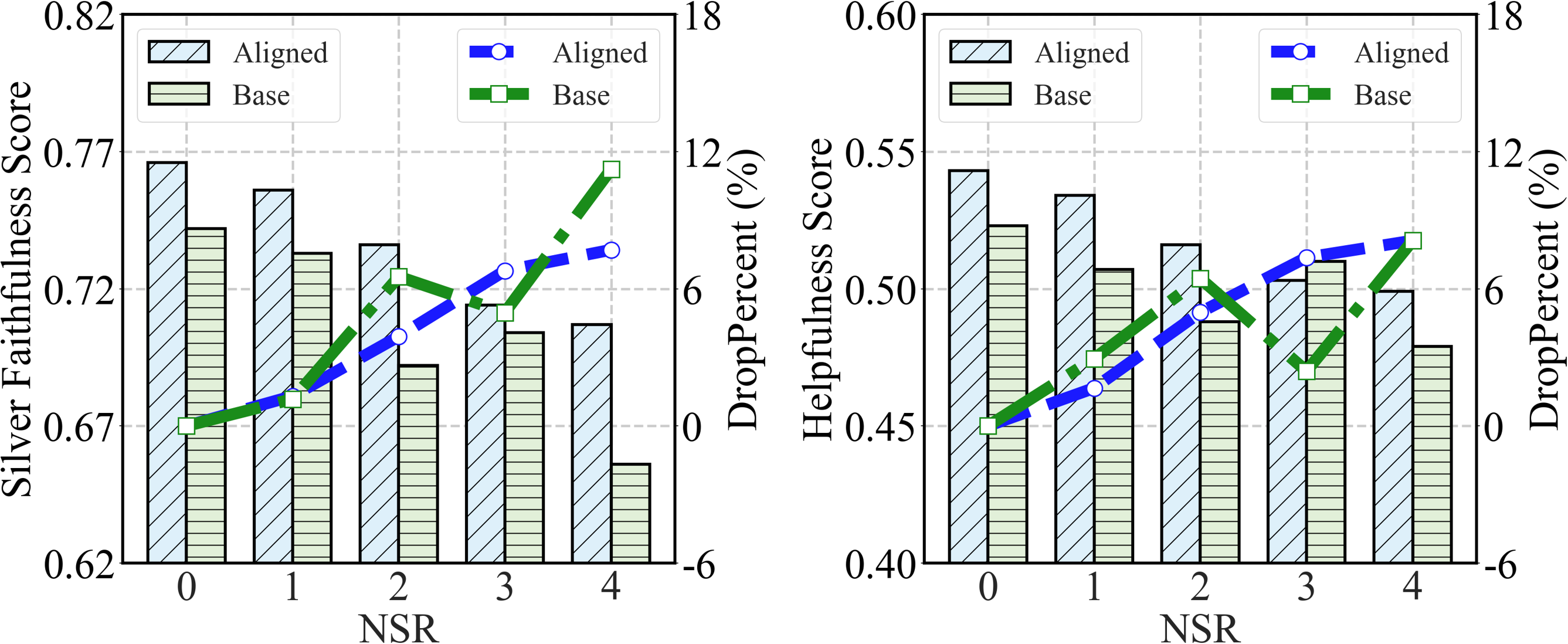}\label{fig:TQA_robust}}
    \subfigure[HotpotQA dataset.]{
        \includegraphics[width=1.0\linewidth]{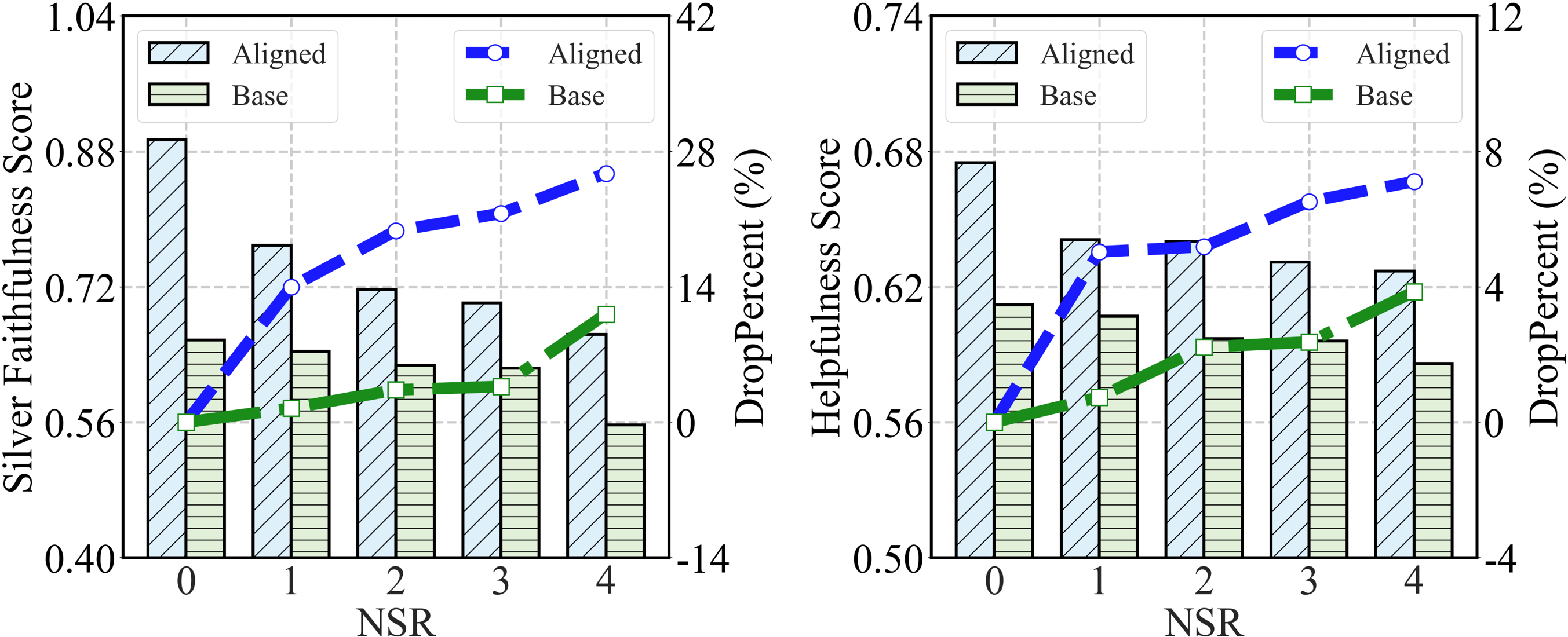}\label{fig:Hot_robust}}
    \caption{Model performance \textit{w.r.t.} Noise-to-Signal Ratio (NSR) ratio. The bar denotes the silver faithfulness score or the helpfulness score, while the line represents the performance drop percent compared to the model that is provided with only relevant retrieved passages.}
    \label{fig:robust_study}
\end{figure}
\subsection{Alignment Study (RQ2)}
To verify the effectiveness of the proposed LPO, we implement \textbf{SEER} with different types of PO methods to optimize the three primary criteria: 
(1) Base, \textit{i.e.,} the base extractor; (2) PPO \cite{schulman2017proximal}; (3) DPO \cite{rafailov2024direct}; (4) LPO (\S\ref{sec:self_alignment}). 
In Figure \ref{fig:align_study}, we present the oracle scores made by each method and the percentage of performance improvement over the Base method. From the results, we find that: 
\textbf{(1)} Unsurprisingly, the Base without alignment performs the worst in 11 out of 12 cases, indicating the necessity of alignment for evidence extraction. 
\textbf{(2)} The PPO usually performs worse than the DPO one, as it directly optimizes the reward signal, \textit{i.e.,} the oracle scores in our work, and thus neglects the pairwise signals between the extracted evidence corresponding to the same query. Besides, the relatively poor performance of PPO may be caused by the difficulty of optimizing PPO, making it hard to reach the optimal point.
\textbf{(3)} Our LPO consistently outperforms the DPO, indicating the superiority of supplementing DPO with a listwise-aware weight. 
\textbf{(4)} After self-alignment, the average improvements of our LPO over the Base on three datasets are 10.2\%, 6.16\%, and 1.70\% regarding the three primary criteria, showing huge potential to enhance the final RAG performance and quicken up the inference.

\subsection{Robustness Analysis (RQ3)}
\label{sec:robustness}
In real-world scenarios, RAG systems usually suffer from data noise issues \cite{DBLP:journals/corr/abs-2312-10997,ding2024survey} caused by imperfect retrieval systems, etc. 
To simulate this scenario, we randomly add a certain proportion (0\%, 100\%, 200\%, 300\%, and 400\%) of irrelevant passages to each test query. We use Noise-to-Signal Ratio (NSR) to denote the ratio of irrelevant passages to the relevant retrieved ones. 
Figure \ref{fig:robust_study} shows the results on silver faithfulness\footnote{The silver faithfulness measures the entailment degree between the relevant retrieved passage (rather than the mixture of it and the irrelevant passages)\, and\, the\, extracted\, evidence.} and helpfulness, while conciseness is omitted as the noise issue does not affect it much. The results show that: 
\textbf{(1)} The performance of both aligned and base extractors decreases, while the aligned one can consistently outperform the base under any NSR except for 1 case. 
\textbf{(2)} The performance drop percent of the aligned model is generally lower than the base in 2 out of 3 datasets. Besides, with 100\% noise proportion, the aligned model can even outperform the base without noise data on all datasets. These observations manifest that \textbf{SEER} can endow the backbone with more robustness to noise issues.

\begin{table}[t]
  \renewcommand\arraystretch{1.5}
  \tabcolsep=0.124cm
  \footnotesize
  \centering
  \begin{tabular}{l|ccc|ccc}
    \toprule
    \multicolumn{1}{c|}{ \multirow{3}{*}[+0.ex]{\centering \bf Model}} & 
    \multicolumn{6}{c}{ \multirow{1}{*}[+0.2ex]{\centering \bf Dataset}} \\
    \cline{2-7}
    
    & \multicolumn{3}{c|}{ \multirow{1}{*}[-0.275ex]{\centering \bf NQ}} & \multicolumn{3}{c}{ \multirow{1}{*}[-0.275ex]{\centering \bf HotpotQA}}\\
    \cline{2-7}
    
    & \multicolumn{1}{c}{ \multirow{1}{*}[-0.275ex]{\centering FS}} & \multicolumn{1}{c}{ \multirow{1}{*}[-0.275ex]{\centering HS}} & \multicolumn{1}{c|}{ \multirow{1}{*}[-0.275ex]{\centering CS}} & \multicolumn{1}{c}{ \multirow{1}{*}[-0.275ex]{\centering FS}} & \multicolumn{1}{c}{ \multirow{1}{*}[-0.275ex]{\centering HS}} & \multicolumn{1}{c}{ \multirow{1}{*}[-0.275ex]{\centering CS}} \\
    \cline{1-7}
    
   \textbf{(A) SEER} & \underline{0.901} & \textbf{0.703} & \underline{0.796} & \underline{0.894} & \textbf{0.674} & \textbf{0.369} \\
   (B) w/o Dup & 0.896 & 0.675 & \textbf{0.800} & 0.881 & 0.657 & \underline{0.365}  \\
   (C) w/o CoV &  \textbf{0.904} & \underline{0.696} & 0.787 & \textbf{0.903} & \underline{0.668} & 0.363  \\
   (D) w/o Lam & 0.894 & 0.684 & 0.785 & 0.857 & 0.654 & 0.359  \\
    \bottomrule
  \end{tabular}
   \caption{Ablation study with key settings of \textbf{SEER}, where we use FS, HS, and CS to indicate the Faithfulness, Helpfulness, and Conciseness scores, respectively.}
  \label{tab:ablation}
\end{table}
\subsection{Ablation Study (RQ4)}
In Table \ref{tab:ablation}, we conduct an ablation study to verify the effectiveness of key settings in our method, where w/o denotes without, 
(A) represents \textbf{SEER}, 
(B) removes the deduplication operation, 
(C) removes smoothing CoV-weighting by uniformly setting $\alpha^f$, $\alpha^h$, and $\alpha^c$ to $1/3$ in Eq. (\ref{eq:cov_score}),
(D) removes the lambda weight $\lambda_{w,l}$ in Eq. (\ref{eq:lpo_loss}). 
From the table, we can find that (A) achieves the best or second-best results in all datasets, indicating all key settings are effective and necessary for \textbf{SEER}. 
By comparing (A) and (B), removing duplicates can considerably improve helpfulness, as it effectively avoids preference optimization overwhelmed by head responses. 
By comparing (A) and (C), weighting the oracle scores based on their statistical properties is able to match the learning difficulty well. 
By comparing (C) and (D), we observe that weighting the preference pairs plays a more key role than weighting the oracle scores. 
The main reason might be that equally treating all preference pairs leads to less attention paid to the crucial ones.

\section{Related works}
\label{sec:related}
\subsection{Context Refinement for RAG}
Recently, many works have emerged, aiming at identifying the supporting content from retrieved passages. The common method is to rerank the retrieved passages and feed the top-ranked ones into generators \cite{zhang2023retrieve,xiao2023c}. 
Thereafter, some methods 
leverage the capabilities of LLMs to summarize retrieved passages to identify key information \cite{ko2024evidence,DBLP:journals/corr/abs-2305-06147,DBLP:journals/corr/abs-2404-13081,DBLP:journals/corr/abs-2401-18059}. 
Furthermore, a few methods leverage agent models to calculate perplexity as an important indicator to filter out low-information content \cite{li2023compressing,DBLP:journals/corr/abs-2310-06839}. 
Other works use manually designed heuristic-based augmentation to construct training signals for fine-tuning LLMs, to enhance their capacity to identify key information \cite{wang2023learning,DBLP:journals/corr/abs-2402-12174}. 
In contrast to previous works heavily relying on hand-crafted augmentation, we 
use data augmented by the model itself to boost performance, free of the arduous workforce.
\subsection{Self-Aligned Learning}
Recently, a few studies have attempted to utilize the model to improve itself and align its response with desired properties \cite{li2023self,zhang2024self,liang2024learning,DBLP:journals/corr/abs-2310-05910,DBLP:conf/icml/YuanPCLSXW24,DBLP:conf/nips/SunSZZCCYG23,DBLP:journals/corr/abs-2212-08073}. 
For example, \cite{li2023self} prompts the model to generate instructions for unlabeled data to create a set of candidate training data, and then use the model to score each augmented example to select high-quality augmented data.
\cite{zhang2024self} utilizes the self-evaluation capability of LLMs to create confidence scores in terms of the factual accuracy of its generated responses, and treat them as reward signals to steer the model towards factuality.
Similarly, \cite{liang2024learning} leverages the model’s self-awareness of its knowledge state to align the model for hallucination mitigation. 
To the best of our knowledge, our study is the first to explore self-aligned learning for evidence extraction.

\vspace{0.1cm}
\section{Conclusion}
\label{sec:conclusion}
This work explores the method that learns to extract high-quality evidence to assist model generation and reduce computational cost. Different from previous works heavily relying on heuristics, we introduce a novel evidence extraction learning framework, \textbf{SEER}, which utilizes the model to calibrate its extraction preference via self-alignment. To this end, we first probe into model extraction preferences via response sampling, then assess the quality of extracted evidence via experts, and finally optimize the vanilla model as an evidence extractor via self-alignment. Extensive experiments show that \textbf{SEER} considerably improves the final RAG performance. Moreover, it can extract more faithful, helpful, and concise evidence, and also shows higher robustness against data noise issues.

\section*{Limitations}
\label{sec:limitation}
Despite our discoveries and improvements, we must acknowledge certain limitations in our work:

\textbf{Firstly}, computing resource constraints restrict our experiment to LLMs with limited and moderate scale, \textit{i.e.,}  Flan-T5-XL \cite{chung2024scaling} and Llama2-7B-Chat \cite{DBLP:journals/corr/abs-2307-09288}. We will explore the use of our method on larger models such as Llama2-70B in future work.
The EM and $\mathrm{F_1}$ metrics used in our experiments might overestimate the correctness of responses, even if the response does not convey equivalent semantics to the ground truth, since these metrics mechanically verify whether the answer exists in the response.

\textbf{Secondly}, our method still requires domain knowledge for devising experts to assess the quality of evidence, though it has considerably lightened the arduous workforce in data engineering. 
We experiment solely on Dense Passage Retriever \cite{karpukhin2020dense} with Wikipedia passages, while de facto RAG applications commonly involve multi-source retrieval with varied writing styles. 
\textbf{Thirdly}, there are a few cases where the aligned extractor is vulnerable to data noise issues. As demonstrated in Figure \ref{fig:Hot_robust}, with the NSR increases, the performance drop percent of the aligned extractor is higher than that of the base one, although it still outperforms the base one. 
Given that, we are currently conducting further research to propose a more powerful evidence extractor, which is not only skilled at refining retrieved passages but also has higher robustness against noisy passages. 

\section*{Acknowledgements}
\label{sec:acknowledgements}
This work is jointly supported by National Natural Science Foundation of China (No. 62376067) and Guangdong Basic and Applied Basic Research Foundation (2023A1515110078).
We sincerely thank all the reviewers for\, their\, detailed\, reviews.

\bibliography{anthology,custom}
\bibliographystyle{acl_natbib}

\appendix

\begin{table}[h]
  \centering
  \footnotesize
  \tabcolsep=0.094cm
  \renewcommand\arraystretch{1.25}
  \begin{tabular}{llcccc}
    \toprule
    \multicolumn{1}{l}{\textbf{Dataset}} & \multicolumn{1}{c}{\textbf{Task}} &  \multicolumn{1}{c}{\textbf{Metric}} & \multicolumn{1}{c}{\textbf{\#Train}} & \multicolumn{1}{c}{\textbf{\#Dev}} & \multicolumn{1}{c}{\textbf{\#Test}}\\
    \midrule 
    NQ & Extractive QA & EM & 79.1k & 8.7k & 3.6k \\
    TQA & Extractive QA & EM & 78.7k & 8.8k & 11.3k \\
    HotpotQA & Abstractive QA & $\mathrm{F}_1$ & 88.9k & 5.6k & 5.6k\\
    \bottomrule
  \end{tabular}
  \caption{\centering{Statistics and task metrics for three datasets.}}
  \label{tab:dataset}
\end{table}

\section{More Implementation Details}
\label{appendix:implementation}
\nosection{Statistics of datasets} We conduct extensive experiments on three benchmark datasets, \textit{i.e.,} NaturalQuestions (NQ) \cite{kwiatkowski2019natural}, TriviaQA (TQA) \cite{joshi2017triviaqa}, and HotpotQA \cite{yang2018hotpotqa}, for evaluating our proposed method and the competitive baselines. 
We show the detailed statistics\, of\, these datasets in Table \ref{tab:dataset}.

\nosection{Response sampling details} 
Given the query and the retrieved passages, we prompt the base extractor to generate 10 candidate response samples and we remove duplicates. 
To fully probe the evidence extraction preferences of the base extractor, we have modified the generation configuration to make the responses more varied. 
Specifically, we set top-p, top-k, temperature, and the repetition penalty as 1.0, 80, 1.0, and 1.0 respectively, for collecting diverse preference data, used to align the responses of the based extractor with\, the\, desired\, properties.  

\nosection{Fine-tuning details} We use the Adam optimizer \cite{DBLP:journals/corr/KingmaB14} with $\beta_1=0.9$, $\beta_2=0.999$, and $eps=1e^{-8}$. 
The learning rate is $1e^{-5}$ with 1.5\% warmup ratio and cosine scheduler. The batch size, gradient accumulation step, and number of epochs are set as 16, 2, and 2.0, respectively.
We leverage the parameter-efficient fine-tuning technique, specifically LoRA \cite{DBLP:conf/iclr/HuSWALWWC22}, where we employ the Llama-Factory\footnote{\url{https://github.com/hiyouga/LLaMA-Factory}.} fine-tuning framework \cite{DBLP:journals/corr/abs-2403-13372} to implement all the preference optimization methods for fair comparisons. 

\nosection{Context relevance details} In Section \ref{sec:prelimi}, we use context relevance as the metric to measure how well the extracted evidence fits the current user query and can be effectively used to augment the quality of generation. 
To this end, we naturally define context relevance as the cosine similarity between the extracted evidence and the user query:
\begin{equation}
     {s}^{cr} = \mathrm{SBERT}_{\mathrm{cosine}}(q,e), 
\end{equation}
where $s^{cr} \in [-1, 1]$ is the context relevance score; $q$ and $e$ denote the query and evidence, respectively.

\nosection{Silver faithfulness details} In Section \ref{sec:robustness}, we devise a metric, silver faithfulness, to measure the robustness of the evidence extractor against data noise issues commonly existing in real-world scenarios.
Specifically, we fed the mixture of the relevant retrieved passage and the randomly sampled irrelevant passages into the extractor.
Then, we treat the relevant retrieved passage and extracted evidence as the premise and hypothesis, respectively, measuring how well the extractor is robust to irrelevant context, which can\, be\, formulated\, as:
\begin{normalsize}
    \begin{equation}
        s^{sf} = \textsc{AlignScore}(\hat{p}, e), \quad e = \tilde{\mathcal{E}}(\cdot|q\oplus \breve{P}),
    \end{equation}
\end{normalsize}where $s^{sf}\in [0, 1]$ is the silver faithfulness score; $\hat{p}$ is the relevant retrieved passage; $\breve{P}$ is the mixture of $\hat{p}$ and those randomly sampled irrelevant\, passages.

\section{Full-length Answer Generation}
\label{appendix:full}
\begin{table*}[t]
    \centering
    \small
    \renewcommand{\arraystretch}{1.4} 
    \begin{tabular}{p{0.965\linewidth}} 
        \toprule
        \textbf{Question:} Which branch of philosophy is concerned with fundamental questions about the nature of reality? \\
        \textbf{Answer:} Metaphysics \\
        \textbf{Full-length answer:} Metaphysics is the branch of philosophy concerned with fundamental questions about the nature of reality. \\
        \midrule \midrule
        \textbf{Question:} What country used the Drachma as its currency, before switching to the Euro in 2001? \\
        \textbf{Answer:} Greece \\
        \textbf{Full-length answer:} Greece used the Drachma as its currency before switching to the Euro in 2001. \\
        \midrule \midrule
        \textbf{Question:} Californian rock band Lit recorded A Place in the Sun in 1995, but what's their best known song? \\
        \textbf{Answer:} My Own Worst Enemy \\
        \textbf{Full-length answer:} The Californian rock band Lit recorded their album A Place in the Sun in 1995, and their best known song is My Own Worst Enemy. \\
        \bottomrule
    \end{tabular}
    \caption{Three examples of full-length answers from the NQ, TQA, as well as HotpotQA\, datasets,\, respectively.}
    \label{tab:full_example}
\end{table*}

\begin{table*}[!h]
    \renewcommand{\arraystretch}{1.4} 
    \begin{tcolorbox}
            \textbf{Full-length Answer Generation Prompt} 
            \tcblower
            \textcolor{black}{\textbf{[Instruction]}}\\
            You are given a question and its answer. Your task is to transform this question-answer pair into a declarative sentence with lossless fidelity to the original semantics. \\
             \textbf{[Here are three examples]} \\
            {[Question]:} What profession does Nicholas Ray and Elia Kazan have in common? \\
            {[Answer]:} director \\
            {[Full-length answer]:}  Nicholas Ray and Elia Kazan have the profession of director in common.\\
            {[Question]:} When is season seven of game of thrones coming out? \\
            {[Answer]:} July 16, 2017 \\
            {[Full-length answer]:} Season seven of Game of Thrones is coming out on July 16, 2017.\\
            {[Question]:} What is the moon festival called in Chinese? \\
            {[Answer]:} Mid-Autumn Festival \\
            {[Full-length answer]:} The moon festival is called the Mid-Autumn Festival in Chinese.\\
            \textbf{[Now complete the following]} \\
            {[Question]:} When did the genre of installation art start to gain acceptance? \\
            {[Answer]:} in the 1970s \\
            {[Full-length answer]:} 
    \end{tcolorbox}
    \caption{The prompt for full-length answer generation.}
    \label{tab:full_prompt}
\end{table*}
To assess the conciseness of the extracted evidence, we propose measuring the information gap between it and the full-length answer. 
The full-length answer is generated by transforming the question and its corresponding answer into a declarative statement, as shown in Table \ref{tab:full_example}. 
Towards this end, we prompt GPT-3.5-turbo to transform each question-answer pair into a full-length answer. Additionally, we prepared a few-shot examples to encourage well-organized output. 
The prompt for full- \quad \quad length answer generation can be found in Table \ref{tab:full_prompt}.

\section{Stability Analysis} 
\label{appendix:stability}
\begin{figure*}[t]
    \centering  
    \subfigure[NQ dataset.]{
        \includegraphics[width=0.32\linewidth]{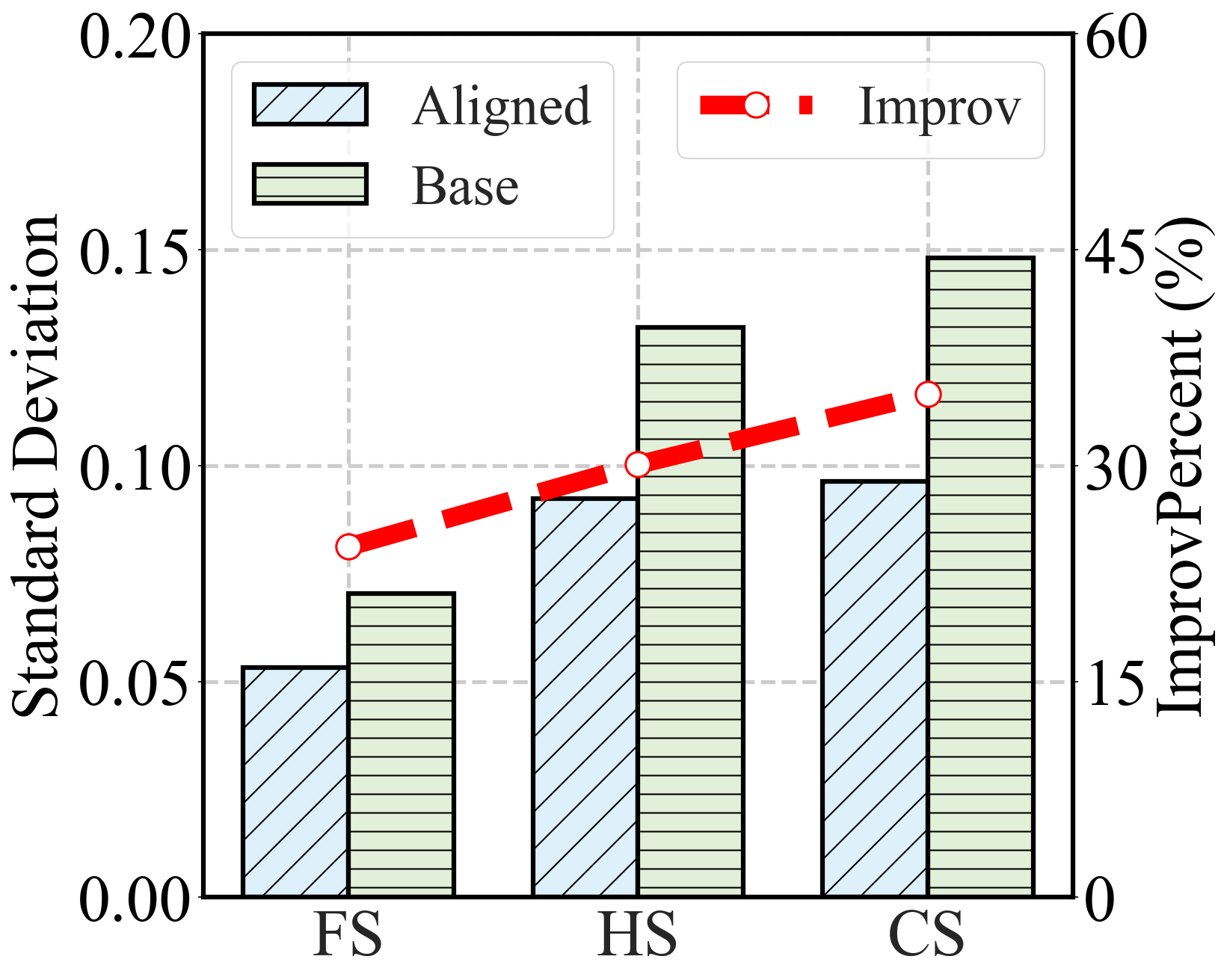}\label{fig:NQ_stability}}
    \subfigure[TQA dataset.]{
        \includegraphics[width=0.32\linewidth]{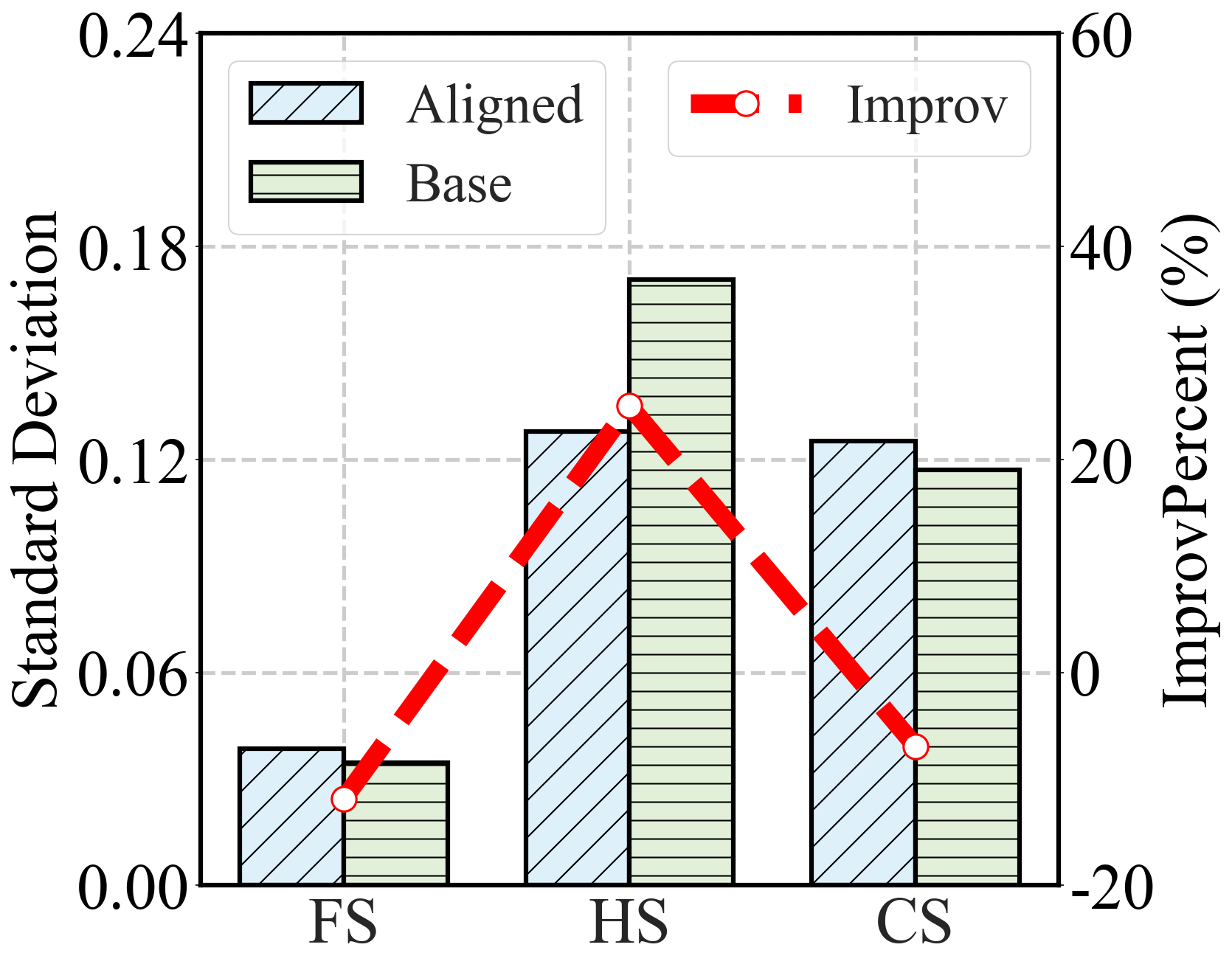}\label{fig:TQA_stability}}
    \subfigure[HotpotQA dataset.]{
        \includegraphics[width=0.32\linewidth]{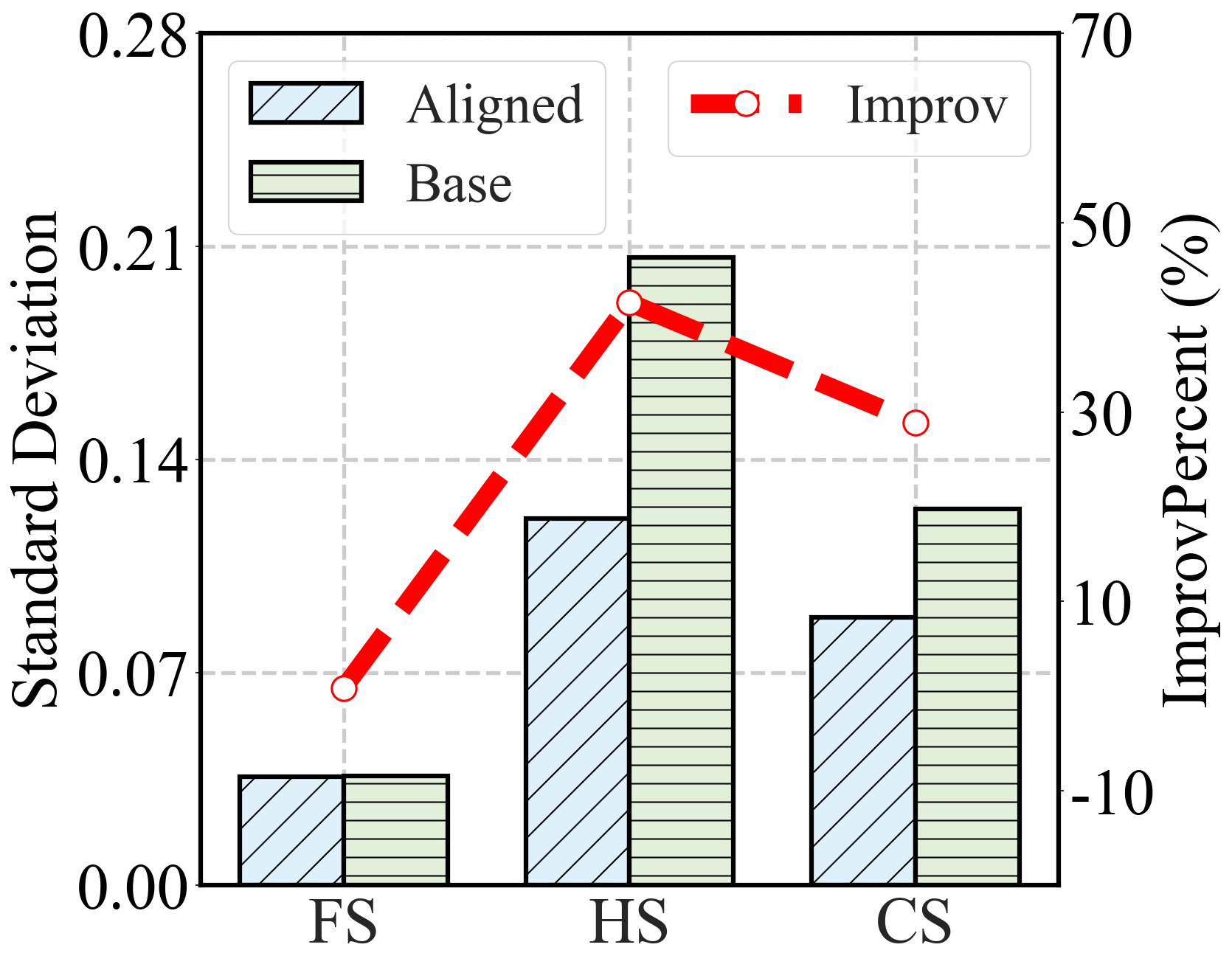}\label{fig:Hot_stability}}
    \vspace{-0.1cm}
    \caption{Model stability \textit{w.r.t.} faithfulness, helpfulness, and conciseness. The bar represents the standard deviation results, while the line represents the stability improvement percent of the aligned model compared to the base model. We use FS, HS, and CS to denote the Faithfulness, Helpfulness, and Conciseness scores, respectively, for simplicity.}
    \label{fig:stability_analysis}
\end{figure*}
In Figure \ref{fig:stability_analysis}, we experiment to verify whether the stability of model generation is improved after self-alignment optimization. 
Specifically, we generate ten pieces of evidence for each test query by response sampling with the same generation configuration as Section \ref{sec:evidence}. 
Subsequently, we measure the oracle scores (\S \ref{sec:expert}), calculate the standard deviation, and compute the average value.
The experimental results show that: 
\textbf{(1)} The generation stability of the aligned model is much better than that of the base one in most cases. More precisely, the average improvement of the aligned model over the base one on the three datasets is 18.5\%. 
\textbf{(2)} The generation stability in terms of helpfulness has seen greater improvements compared to the other two properties (\textit{i.e.,} faithfulness and conciseness), with an average improvement of 32.2\%, showing the huge potential to enhance the final RAG performance. 
The above observations fully demonstrate that \textbf{SEER} is able to endow the backbone with superior generation stability during the inference.

\section{Learning Algorithm of SEER} 
\label{appendix:seer_alg}
Algorithm \ref{alg:seer_alg} demonstrates the learning algorithm of the proposed \textbf{SEER} framework. The algorithm can be divided into three stages, \textit{i.e.,} 
\textbf{(1) Evidence Extraction} (line 3-6), \textbf{(2) Expert Assessment} (line 7-10), as well as \textbf{(3) Self-Alignment} (line 11-14).
\begin{algorithm*}[!h] 
    \renewcommand{\arraystretch}{1.5} 
    \caption{Learning algorithm of \textbf{SEER}}
    \label{alg:seer_alg}
    \renewcommand{\algorithmicrequire}{\textbf{Input:}}
    \renewcommand{\algorithmicensure}{\textbf{Output:}}
    \begin{algorithmic}[1]
        \REQUIRE Trainig dataset with queries $q$, answers $a$, and retrieved passages $P=\{p_i\}_{i=1}^K$; the base evidence extractor $\mathcal{E}$; the sample size $M$; total number of iterations $T$. 
        \ENSURE The aligned evidence extractor $\tilde{\mathcal{E}}$    
        
        \STATE Initialize the model parameter $\tilde{\mathcal{E}}$ with $\mathcal{E}$
        \FOR{each $i \in [1, T]$}
            \STATE \# \textbf{Stage1: Evidence Extraction}   
            \STATE Sample a mini-batch of ($q$, $a$, $P$) query-answer-passage triples from the dataset.
            \STATE Get evidence candidates $\{e_j\}_{j=1}^M$ via response sampling $e \sim \mathcal{E}(\cdot|q\oplus P)$.
            \STATE Obtain uniformly distributed set $\{e_j\}_{j=1}^N$ by removing duplicates in $\{e_j\}_{j=1}^M$.
            
            \STATE \# \textbf{Stage2: Expert Assessment}  
            \STATE Construct a QuadQARE for each evidence candidate $<q, a, P, e>$.
            \STATE Get the oracle scores $(s^f, s^h, s^c)$ for each evidence candidate with Eq. (\ref{eq:faith_score}-\ref{eq:concise_score}).
            \STATE Get the smoothing CoV-weighted score $s$ with Eq. (\ref{eq:sigma_mu}-\ref{eq:cov_score}).

            \STATE \# \textbf{Stage3: Self-Alignment}  
            \STATE Get the lambda weight $\lambda_{w,l}$ for each preference pair $(x, y_w, y_l)$ with Eq. (\ref{eq:lambda_weight}).
            \STATE Compute the preference optimization loss $\mathcal{L}_{\mathrm{LPO}}$ with Eq. (\ref{eq:lpo_loss}).
            \STATE Update the model parameter of $\tilde{\mathcal{E}}$ using gradient descent.
        \ENDFOR
        \RETURN $\tilde{\mathcal{E}}$.
    \end{algorithmic}
\end{algorithm*}

\end{document}